\documentclass[letterpaper]{article} 
\usepackage{aaai24}  
\usepackage{times}  
\usepackage{helvet}  
\usepackage{courier}  
\usepackage[hyphens]{url}  
\usepackage{graphicx} 
\usepackage{natbib}  
\usepackage{caption} 
\usepackage{algorithm}
\usepackage{algorithmic}
\usepackage{amssymb}
\usepackage{multirow}
\usepackage{bm}
\usepackage{subfigure}
\usepackage{array}

\newtheorem{definition}{Definition}

\newtheorem{theorem}{Theorem}
\newtheorem{assumption}{Assumption}
\newtheorem{corollary}{Corollary}

\usepackage{newfloat}
\usepackage{listings}
\usepackage{amsmath}
\usepackage{color}
\usepackage{newfloat}
\usepackage{listings}
\DeclareCaptionStyle{ruled}{labelfont=normalfont,labelsep=colon,strut=off} 
\floatstyle{ruled}
\newfloat{listing}{tb}{lst}{}
\floatname{listing}{Listing}
\title{Rethinking Causal Relationships Learning in Graph Neural Networks}
\author{
    Hang Gao \textsuperscript{\rm 1, \rm 2}\equalcontrib,
    Chengyu Yao\textsuperscript{\rm 1, \rm 2}\equalcontrib,
    Jiangmeng Li \textsuperscript{\rm 1, \rm 3},
    Lingyu Si \textsuperscript{\rm 1, \rm 2},
    Yifan Jin \textsuperscript{\rm 1, \rm 2},
    Fengge Wu \textsuperscript{\rm 1, \rm 2}\thanks{Corresponding authors.},
    Changwen Zheng \textsuperscript{\rm 1, \rm 2}\footnotemark[2],
    Huaping Liu \textsuperscript{\rm 4}
}

\affiliations{
    \textsuperscript{\rm 1}Science and Technology on Integrated Information System Laboratory, \\Institute of Software Chinese Academy of Sciences\\
    \textsuperscript{\rm 2}University of Chinese Academy of Sciences\\
    \textsuperscript{\rm 3}State Key Laboratory of Intelligent Game\\
    \textsuperscript{\rm 4}Department of Computer Science and Technology, Tsinghua University\\
    \{gaohang, yaochengyu2023, jiangmeng2019, lingyu, yifan2020, changwen, fengge\}@iscas.ac.cn, \\
    hpliu@tsinghua.edu.cn
}

\usepackage{bibentry}

\begin{document}

\maketitle

\begin{abstract}
Graph Neural Networks (GNNs) demonstrate their significance by effectively modeling complex interrelationships within graph-structured data. To enhance the credibility and robustness of GNNs, it becomes exceptionally crucial to bolster their ability to capture causal relationships. However, despite recent advancements that have indeed strengthened GNNs from a causal learning perspective, conducting an in-depth analysis specifically targeting the causal modeling prowess of GNNs remains an unresolved issue. In order to comprehensively analyze various GNN models from a causal learning perspective, we constructed an artificially synthesized dataset with known and controllable causal relationships between data and labels. The rationality of the generated data is further ensured through theoretical foundations. Drawing insights from analyses conducted using our dataset, we introduce a lightweight and highly adaptable GNN module designed to strengthen GNNs' causal learning capabilities across a diverse range of tasks. Through a series of experiments conducted on both synthetic datasets and other real-world datasets, we empirically validate the effectiveness of the proposed module. The codes are available at \url{https://github.com/yaoyao-yaoyao-cell/CRCG}.
\end{abstract}

\section{Introduction}

\begin{figure}[h]
	\centering
	\subfigure[The performance of DIR \cite{DBLP:conf/iclr/WuWZ0C22}, a GNN with causal enhancement method, and Empirical Risk Minimization (ERM), as conventional GNN, on datasets devoid of confounders. ERM and DIR employ identical backbone architectures with consistent network sizes.]{
		\begin{minipage}{0.45\textwidth} 
			\includegraphics[width=\textwidth]{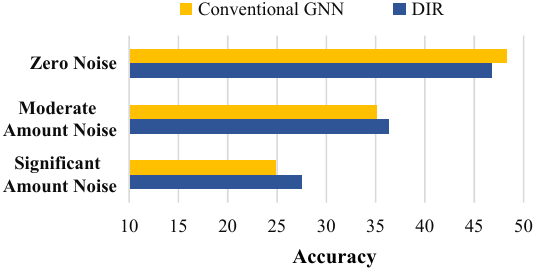} 
		\end{minipage}
	}
	\subfigure[The performance improvement achieved by DIR on the GNN model when confronted with varying degrees of correlation between confounders and causal factors. The degree of correlation denotes the increment in the probability of the corresponding confounder appearing when specific causal factors are present, as compared to the probability of other random noise occurrences. Therefore, the larger degree of correlation is, the stronger the interference of the confounder.]{
		\begin{minipage}{0.45\textwidth}
			\includegraphics[width=\textwidth]{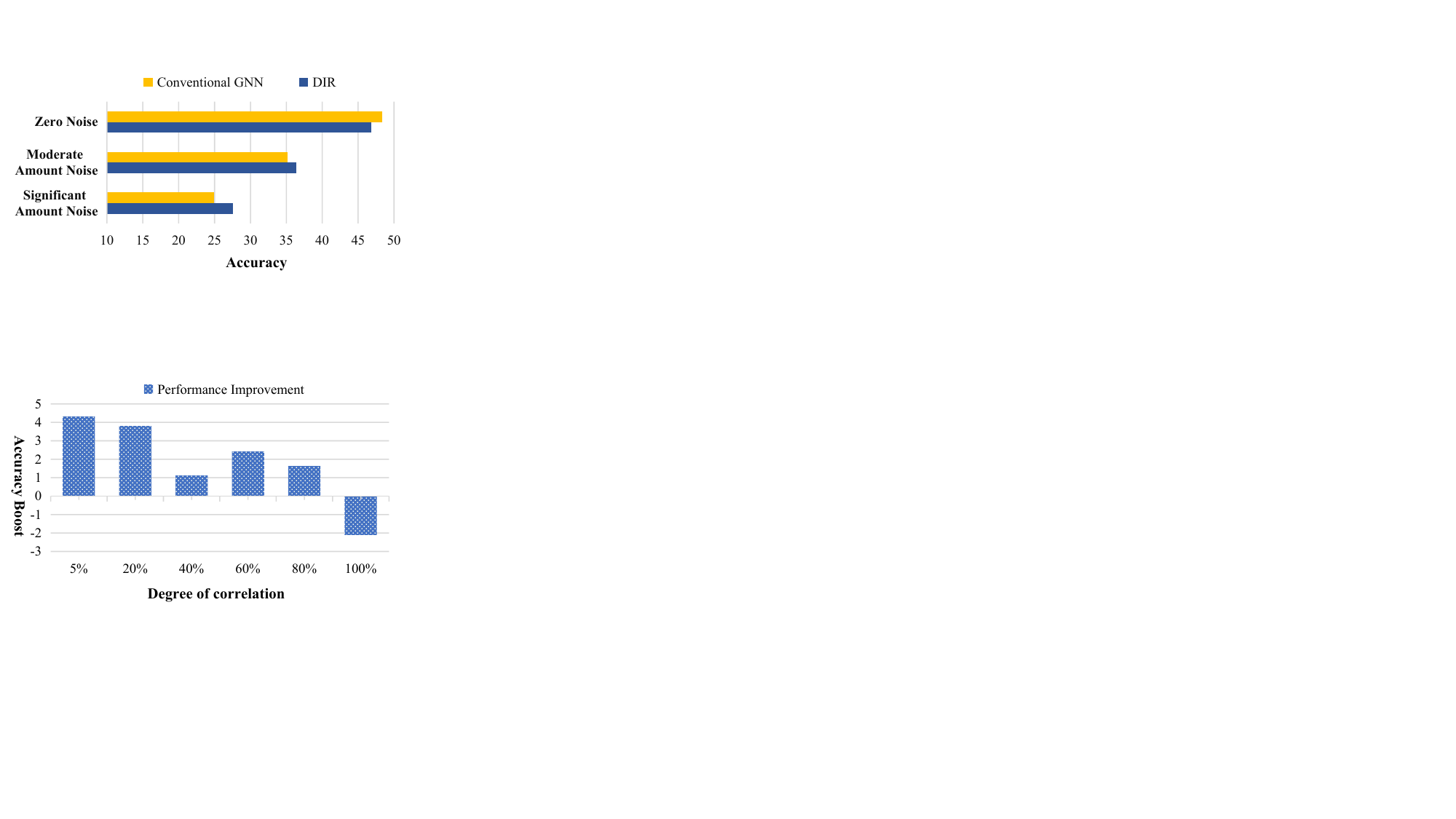}
		\end{minipage}
    }
	
	\caption{Experimental results with CRCG.}
	\label{fig:mtv}
\end{figure}

Graph representation learning is a fundamental challenge across diverse domains. It involves mapping intricate graphs into compact vector representations while retaining vital structural and semantic insights. By incorporating neural networks, GNNs \cite{DBLP:journals/access/BilotMAZ23} have emerged as potent tools for addressing such a challenge. However, GNNs typically model statistical, not causal, relationships between data and labels. This can compromise reliability, especially with intricate graph data. Recognizing this, there's a growing emphasis on enhancing GNNs' causal modeling capabilities. Enabling GNNs to grasp causal links between data and labels can bolster robustness and credibility, leading to superior outcomes in various real-world scenarios.

Currently, there are several emerging approaches aimed at enhancing the causal modeling capability of GNNs while maintaining an end-to-end framework. These methods aim to eliminate the influence of confounder within graph data, as confounder can create a false association between the cause and effect due to its correlation with both \cite{DBLP:journals/aim/Pearl02}. Specifically, some approaches \cite{DBLP:conf/iclr/WuWZ0C22,DBLP:conf/nips/Fan0MST22} involve partitioning the training data into causal components and confounders, followed by separate processing to enable the model to disregard the confounders. Alternatively, other methods \cite{DBLP:conf/nips/0002ZB00XL0C22,DBLP:conf/aaai/GaoLQSXZ023} aim to directly identify causal data or eliminate confounders to achieve the modeling of causal relationships. Furthermore, there exists a multitude of techniques that center their focus on studying the modeling capability of GNNs for specific causal relationships in practical application scenarios \cite{DBLP:journals/www/CaoHRXXA23, DBLP:journals/concurrency/GaoLW22, DBLP:conf/aaai/0001ACSVM22}. These GNN causal enhancement methods have all demonstrated favorable outcomes, effectively enhancing the robustness and credibility of GNN models. Furthermore, these approaches don't alter the network backbone; rather, they introduce new modules or adjust training processes to enhance causality. While relevant models have achieved some progress in enhancing the modeling of causal relationships within the GNNs, there is still a lack of in-depth research in this area.

To address such an issue, we aim at conducting a comprehensive and detailed analysis. The analysis starts by studying the dataset and observing how confounders in the data might impact GNN training. However, due to the complexity of graph data, manually identifying such confounders and their specific effects is challenging. Thus, we constructed a synthetic dataset called \textbf{C}ausal \textbf{R}elationship \textbf{C}onfigurable \textbf{G}raph (CRCG) dataset, which can generate complex graph data with explicitly identifiable and controllable causal relationships. We have also theoretically demonstrated the rationality of the data generation process for the CRCG dataset.

Utilizing the CRCG dataset, we conducted a series of experiments to compare the performance differences between GNN with causal enhancement and conventional GNN under different scenarios. Figure \ref{fig:mtv} presents the results. It is evident that in the presence of confounders within graph data, the GNN with causal enhancement method does exhibit a certain degree of effectiveness. However, in scenarios without confounder, conventional GNN performs on par or even outperforms GNN with causal enhancement. Additionally, we observed that as the correlation degree between the confounder and causal factor changes, the advantage of GNN with causal enhancement diminishes. Further experimental results in Table \ref{tab:nco} and \ref{tab:t2} also suggest the same phenomenon on more baselines. The experimental findings reveal that GNNs with causal enhancement did not succeed in completely eliminating confounder across all scenarios. Moreover, in scenarios without confounder, they could even have a counterproductive effect.

To explain this experimental phenomenon, we conducted a more in-depth analysis, both theoretically and empirically, leading to a conclusion. It states that current GNN causal enhancement methods essentially manipulate the GNNs by applying operations based on certain priors to mitigate the impact of confounding factors on the model's outputs. And such operation needs to conduct with a prior of the graph data.
Furthermore, such interventions can be affected by changes in the dataset, particularly the probabilistic correlation between confounders and causal elements. Building upon the aforementioned findings, we propose that since the primary objective is to minimize the influence of confounders on the model's outputs, it is sufficient to apply operations directly to the model's output representations. This approach reduces the need for introducing additional neural networks, thereby simplifying the model. Furthermore, we can make the model more flexible and adjustable to accommodate various datasets.

Based on this line of thought, we introduce a lightweight module called the Representation-based Causality Augmentation Module (R-CAM) to optimize the GNNs' ability in modeling causal relationships. R-CAM operates in a plug-and-play manner and can be seamlessly applied to various GNN models. R-CAM compels GNN models to acquire more causal knowledge by accentuating features causally linked to labels and disregarding features devoid of causal relationships with the labels. The introduced prior knowledge in R-CAM can be easily tailored to suit different datasets. Our multiple experiments on both artificially synthesized datasets and real-world datasets have demonstrated the efficacy of R-CAM.

Our contributions are as follows:
\begin{itemize}
\item We construct a novel synthetic graph dataset, CRCG, with inherent causal relationships and controllability. CRCG significantly surpasses existing datasets of similar nature. Additionally, the rationality of the data generation process for CRCG has been theoretically demonstrated.

\item We conducted an array of analyses on various GNN models using the CRCG dataset and arrived at corresponding conclusions. Both theoretical and experimental evidence substantiates our findings.


\item Building upon our findings, we devise a novel plug-and-play module named R-CAM. R-CAM is applicable across various GNN models and enhances their capacity for causal relationship modeling. Through experiments conducted on both artificially synthesized and real-world datasets, we validate the efficacy of R-CAM.

\end{itemize}

\begin{table*}[ht]\small
	\setlength{\tabcolsep}{5pt}
 	\caption{Comparative analysis of our dataset with other similar datasets. The term ``Total Combination'' refers to the maximum possible number of combinations attainable when all available graphical elements are employed and juxtaposed in pairs. Please refer to \textbf{Appendix} B for further details.}
	\label{tab:nb}
	\begin{center}
		\begin{tabular}{lcccccc}
			\hline\rule{-3pt}{10pt}

			\multirow{2}*{Dataset}    & Subgraph & Adjustable &Concatenation & Node Feature  &Adjustable    & Total   \\
                 & Types &Subgraph Shapes &Methods  &Generation Methods &Feature Generation   &  Combinations  \\
			\hline\rule{-3pt}{10pt}
			\text{Synthetic Graph}  & \multirow{2}*{5} & Partially & \multirow{2}*{1} & \multirow{2}*{1}& Not  & \multirow{2}*{25}\\
   			\cite{DBLP:conf/nips/YingBYZL19}  &  & Adjustable &  & & Adjustable & \\
   
   			\text{Spurious-Motif }  & \multirow{2}*{6} & Partially & \multirow{2}*{1} & \multirow{2}*{1}& Not & \multirow{2}*{36}\\
         	\cite{DBLP:conf/iclr/WuWZ0C22}  &  & Adjustable &  & & Adjustable & \\
      
            \hline \rule{-3pt}{10pt}
			\multirow{2}*{CRCG}  & \multirow{2}*{25} & Fully & \multirow{2}*{4} &\multirow{2}*{25}  &  Fully & \multirow{2}*{3750}  \\
   			  & & Adjustable &  &  &  Adjustable & \\
			\hline 
		\end{tabular}
	\end{center}


\end{table*}

\section{Related Works}
\subsection{Graph Neural Networks}
GNNs have garnered significant attention in recent years due to their remarkable capability in learning from graph-structured data. Early GNNs laid the foundation for node-level and graph-level representation learning \cite{DBLP:conf/iclr/KipfW17, DBLP:conf/iclr/VelickovicCCRLB18, DBLP:conf/iclr/XuHLJ19}. Since then, a multitude of GNN variants have emerged \cite{DBLP:conf/iclr/WangYZ022, DBLP:conf/icml/FuZB22, DBLP:conf/iclr/ZhangLSS22}, each addressing specific challenges. In addition, there's an increasing interest in enhancing GNNs' ability to model causal relationships \cite{DBLP:conf/iclr/WuWZ0C22}, as GNNs with causal enhancement aim to incorporate causal inference into graph learning, leading to more reliable predictions. 



\subsection{Causal Learning}
Causal learning aims at inferring and understanding causal relationships between events. Current causal learning can be divided into two main directions: causal inference and causal discovery \cite{zhou2018causal, athey2018estimation, cheng2019nonparametric}. The optimization of neural network robustness and reliability through causal learning methods has emerged as a focal point of research interest among scholars \cite{DBLP:conf/nips/LiQZMZ0X22, DBLP:conf/icml/AhujaMWB23, DBLP:conf/coling/JinLLJH22}. Recently, causal learning methods have also been widely used in graph neural networks . Such methods \cite{DBLP:conf/iclr/WuWZ0C22,DBLP:conf/nips/0002ZB00XL0C22,DBLP:conf/aaai/GaoLQSXZ023} discovered potential laws in graph representation learning by studying causality in graph learning, and improved the completion effect of corresponding downstream tasks. We aim to develop a sound analytical approach to thoroughly analyze these methods.

\section{Evaluation on the Causal Modeling Capability of GNNs}
\subsection{Preliminaries}
\subsubsection{Causal Model}
In the realm of causality \cite{pearl2000models}, researchers analyze causal relationships within a system by employing causal models. A causal model is a framework used to represent the causal relationships between different variables or factors in a system. A causal model $\mathcal{M}$ can be represented as a graph, where variables are connected by directed edges to indicate the direction of influence. For a variable $X$, its ancestor $S$ in a causal model is a variable that directly or indirectly influences $X$. On the other hand, descendant $D$ is a variable that is directly or indirectly influenced by $X$. In our analysis, we assume the existence of a causal model $\mathcal{M}$ that can be used to model our task. However, the specific structure of this model is currently unknown to us.

\subsection{CRCG Dataset}
Firstly, we present the details of our proposed CRCG dataset. To thoroughly analyze the ability of GNNs in modeling causal relationships from multiple perspectives, the CRCG dataset is created as a synthetically generated dataset that allows for the construction of various causal relationships as needed. Table \ref{tab:nb} gives a comparison of CRCG with other synthetic graph datasets with controllable causal relationships. The CRCG dataset is designed to create graphs with intricate structures and node features. It involves utilizing various controllable subgraphs to construct the entire graph through distinct connection methods. Node features are also generated using diverse patterns. A detailed description of the dataset can be found in \textbf{Appendix} B. 

CRCG dataset offers a more diverse and intricate set of graph data to enable rigorous testing of GNNs in more complex scenarios. Not only does the CRCG dataset provide a wider range of graph data construction patterns, but it also allows for the adjustment of these patterns through parameters, significantly enriching the foundational dataset for analyzing graph learning algorithms. Furthermore, despite generating a large number of complex graph data, the entire data generation process of CRCG is pre-known and understood, facilitating causal analysis of neural networks trained on this dataset.


\subsubsection{Data Generation.}
We now proceed with an analysis to understand how to effectively generate data based on CRCG. The graph data $G$ can be decomposed into three components: the causal factors $X$ that have a causal relationship with the labels, the confounder $C$ that are probabilistically related to the labels but lack a causal relationship, and the purely independent noise components $U$. Modeling $X$ and $U$ is relatively straightforward within our dataset since we can determine the labels based on $X$ and add randomly generated noise data as $U$. However, establishing $C$ as a variable that complicates and challenges the modeling of causal relationships requires more rigorous theoretical guidance. We employ the following theorem to guide the construction process of $C$.
\begin{theorem}
Assuming that the generation process of the graph data $G$ follows a Markov process, then the set of confounders $C$ in $G$ must be descendants of the set of causal factors $X$ in $G$ or their ancestors. 
\label{lma:da}
\end{theorem}
The proof can be found in \textbf{Appendix} A.1. Due to the fact that our dataset is constructed based on a series of decisions and computations, the data generation process conforms to a Markov process. Hence, we adhere to Theorem \ref{lma:da} to generate the confounder $C$. Specifically, given the manipulability of the data generation process for CRCG, our objective is to ensure that certain aspects of confounders are determined by specific causal factors, as opposed to random data.


\subsection{Evaluations}

Drawing on the CRCG dataset and relevant theories, we have the capacity to conduct both theoretical and empirical analyses concerning the causal modeling capabilities of diverse GNN models. In the domain of causality research, the causal impact of variable $X$ on variable $Y$ can be effectively expressed through the causal effect $P(Y|\hat{X})$ \cite{pearl2000models}, with $\hat{X}$ representing the intervention operation on variable $X$. However, intervention operations necessitate data manipulation, value assignment, and observation of corresponding responses, which is challenging to achieve within the training context of GNNs. In order to analyze the causality of knowledge acquired by GNN models, we propose a novel concept  ``causally estimability,'' and employ it as a criterion for assessing the causal learning capabilities of GNN models.
\begin{definition} (Causally Estimability)
Assuming there exists a GNN $f_{\theta^{*}}(\cdot)$ that models the causal effect $P(Y|\hat{G})$, then the causal effect $P(Y|\hat{G})$ is said to be ``causally estimable'' if the following equation holds:
\begin{gather}
	 \theta^{*} = \arg\min_{\theta}\Big( \sum_{i=1}^{n} \mathcal{H}\big(f_{\theta}(G_{i}),Y_{i}\big)\Big),
	\label{eq:lr}
\end{gather}
where $G_{i}$ is a graph sampled from the value space $\mathcal{G}$ of $G$.  $Y_{i}$ denotes the corresponding ground-truth label. $n$ is the number of sampled graph data with a sufficiently large value. $\mathcal{H}$ denotes the cross-entropy loss. $f(\cdot)$ denotes a GNN that models probabilistic relation between $G$ and $Y$. $\theta$ and $\theta^{*}$ denotes the network parameters of $f(\cdot)$.
\label{def:ce}
\end{definition}
Definition \ref{def:ce} provides a precise framework for modeling causal effects within the realm of graph representation learning. The underlying concept of this definition is notably intuitive. Drawing inspiration from \cite{DBLP:conf/kdd/Pearl11}, we can view a causal relationship as a theorem that can be formalized as a function. Consequently, base on Universal Approximation Theorem \cite{cybenko1989approximation}, if causal effects can be accurately manifested within the data, they can be effectively approximated through training—a quality we term as causally estimability.

Next, we proceed to analyze the relationship between $Y$ and $G$. Within the CRCG dataset, all labels can be determined based on the information within the graph data $G$. And, in real-world scenarios, graph data labels are typically annotated based on the content of the data. Therefore, we can actually consider that $G$ truncates the influence of all its ancestors on Y. To facilitate subsequent analysis and reduce unnecessary interference, we propose the following assumptions. 
\begin{assumption}
For any ancestor $S$ of $G$, the conditional independence $S \perp \!\!\!\perp Y | G$ holds.
\label{asp:idn}
\end{assumption}
With Definition \ref{def:ce} and Assumption \ref{asp:idn}, we can analyze the model's ability to model causal relationships under the absence of confounders $C$. Theoretically, we propose the following theorem.
\begin{theorem}
If there are no confounders in $G$, and Assumption \ref{asp:idn} holds, it can be asserted that the causal effect $P(Y|\hat{G})$ is causally estimable.
\label{thm:noc}
\end{theorem}
The proof can be found in \textbf{Appendix} A.2. Theorem \ref{thm:noc} suggests that if the model's expressive capacity is sufficiently strong to model specific causal relationships, and there are no confounders present in the data, then the said causal relationships are causally estimable. However, in practical scenarios, even in the absence of confounders, the complexity of the dataset can still introduce interference. We will conduct experimental analysis on a dataset without confounders to compare the performance of conventional GNNs with GNNs with causal enhancement modules.


We adopt ERM and ASAP \cite{ranjan2020asap} as foundational benchmarks for conventional GNNs. Additionally, for GNNs with causal enhancement, we pick DIR \cite{DBLP:conf/iclr/WuWZ0C22}, CIGA \cite{DBLP:conf/nips/0002ZB00XL0C22}, DISC \cite{fan2022debiasing} and RCGRL \cite{DBLP:conf/aaai/GaoLQSXZ023} as our baseline methods. These methods adopt the same GNN backbone as ERM. The details of the methods can be found in \textbf{Appendix} C.1. We first test the baselines under the scenario with no confounders. We utilize our proposed CRCG to generate the corresponding data. The details of the experiment settings and dataset can be found in \textbf{Appendix} C.2 and C.3.

\begin{table}[ht]\scriptsize
	\setlength{\tabcolsep}{5pt}
 	\caption{Performance of different baselines on the dataset without confounder. The numbers in ``( )'' indicate relative performance compared to ERM: \textcolor{green}{green} for improvement, \textcolor{red}{red} for inferiority.}
	\label{tab:nco}
	\begin{center}
		\begin{tabular}{l|cccccc}
			\hline\rule{-3pt}{10pt}

			Method  &  noise=0  & noise=1  & noise=2   \\
			\hline\rule{-3pt}{10pt}
			\text{ERM} & 48.33$\pm$0.70 & 35.16$\pm$1.42 & 24.91$\pm$1.21 \\
   			\text{ASAP} & 48.94$\pm$0.63 (\textcolor{green}{+0.61}) & 33.52$\pm$1.34 (\textcolor{red}{-1.64}) & 26.35$\pm$0.88 (\textcolor{green}{+1.44}) \\
      \hline
      \hline\rule{-3pt}{10pt}
			\text{DIR } & 46.80 $\pm$0.92(\textcolor{red}{-1.53}) & 36.37$\pm$1.18(\textcolor{green}{+1.21}) & 27.53$\pm$1.02(\textcolor{green}{+2.62}) \\
			\text{CIGA } & 43.18$\pm$1.24(\textcolor{red}{-5.15}) & 26.42$\pm$1.38(\textcolor{red}{-8.74}) & 24.47$\pm$1.29(\textcolor{red}{-0.44})  \\
			\text{RCGRL } & 52.72$\pm$1.60(\textcolor{green}{+4.39}) & 30.50$\pm$0.52(\textcolor{red}{-4.66}) &  26.44$\pm$1.26(\textcolor{green}{+1.53})  \\
   			\text{DISC } & 45.60$\pm$0.79(\textcolor{red}{-2.73}) & 38.35$\pm$1.31(\textcolor{green}{+3.19}) &  26.80$\pm$0.98(\textcolor{green}{+1.89})  \\
			\hline 
		\end{tabular}
	\end{center}
\end{table}

Results in Table \ref{tab:nco} show that, like in the introduction, methods other than DIR face similar situations. Both GNNs with causal enhancement and regular GNNs perform similarly, lacking a clear edge. Sometimes, GNNs with causal enhancement even perform worse. Given Theorem \ref{thm:noc}, GNNs can model causal relationships in confounder-free graph data, but current GNNs fall short due to limited capabilities. In other words, the model's success depends on the GNN's ability to capture data's probabilistic relationships. This explains why GNNs with causal enhancement don't excel on this dataset. However, questions remain: why do GNNs with causal enhancement sometimes lag behind regular GNNs? And why don't they consistently outperform when dealing with datasets containing confounders? To address these questions, further analysis and experiments are required.

Therefore, we first conducted a theoretical analysis of the model's ability to capture causal relationships on datasets containing confounders. The following theorem encapsulates our conclusions.
\begin{theorem}
If Assumption \ref{asp:idn} is satisfied and a confounder $C$ exists within graph $G$, then $P(Y|\hat{G})$ is not causally estimable. However, such estimation becomes attainable if an intervention $do(C)=\widetilde{C}$ is feasible for all $\widetilde{C} \in \mathcal{C}$, where $\mathcal{C}$ denotes the value space encompassing all potential values of $C$.
\label{thm:do}
\end{theorem} 
The proof can be found in \textbf{Appendix} A.3. Intervention $do(\cdot)$ denotes an operation that deliberately alters or modifies a factor in a system to observe its impact on other variables. Theorem \ref{thm:do} presents a framework for mitigating confounders in the learning process of GNNs. Another perspective on intervention, as posited by \cite{pearl2000models}, involves treating the force responsible for the intervention as a variable. We extend this notion to denote any operation that may impact the training procedure of GNN as variables, thus broadening the utility of Theorem \ref{thm:do} for diverse methodological investigations. Specifically, we present the following corollary.
\begin{corollary}
Under the conditions specified in Theorem \ref{thm:do}, if there exists an operation $T$ such that $f(G) \perp \!\!\! \perp C \ | \ T$ and $I\big((f(G);X \ | \ T\big) = I\big(f(G);X\big)$, then the causal estimability of $P(Y|\hat{G})$ is guaranteed given such $T$.
\label{cly:cd}
\end{corollary} 
Proof can be found in \textbf{Appendix} A.4. Corollary \ref{cly:cd} states any operation can substitute Theorem \ref{thm:do}'s intervention, given Corollary \ref{cly:cd}'s conditions met. This broader characterization can be used to effectively describe the existing GNNs with causal enhancement, as they essentially rely on adopting certain operations to mitigate confounders. However, as the operation must satisfy $f(G) \perp \!\!\! \perp C \ | \ T$ and $I\big((f(G);X \ | \ T\big) = I\big(f(G);X\big)$, shifts in dataset distribution can lower its efficacy. This explains earlier GNNs with causal enhancement's reduced performance on generated datasets.

Next, we conducted three distinct types of experiments to empirically analyze the impact of confounders. Firstly, from a probabilistic perspective, we adjusted the magnitude of the confounder. Based on Theorem \ref{lma:da}, within the training set, we establish causal relationships between the confounders and the causal factors with varying probabilities $P$. In the testing set, we remove such relationships to assess whether the GNN model is influenced by the confounders. The details of the experiment settings and dataset can be found in \textbf{Appendix} C.2 and C.3.  

The experimental results are demonstrated in Table \ref{tab:t2}. From the results, we can observe that as $P$ varies, the advantage of causal GNN over a conventional GNN gradually shifts. This indicates that, in practice, GNNs with causal enhancement might not effectively eliminate confounders in all scenarios; instead, they can yield favorable outcomes only in certain cases.

For further analysis, we conduct two additional experiments. One with changing size of confounders, the other with changing complexity relation between confounder and causal factors. The details of the experiment settings and dataset can be found in \textbf{Appendix} C.2 and C.3. The experimental results are demonstrated in Table \ref{tab:t3} and \ref{tab:t4}. We can observe from the results that, although the performance improvement offered by various causal-enhanced GNN algorithms, as compared to conventional GNNs, does experience certain adjustments with variations in the size of the confounder and the intricacy of its connection with the causal factor, these adjustments are not as significant as the ones seen in Table \ref{tab:t2}. This suggests that the probabilistic relationship between the confounder and the causal factor is the primary factor influencing the effectiveness of causal-enhanced GNN algorithms.

\begin{table*}[ht]\scriptsize
	\setlength{\tabcolsep}{10pt}
 	\caption{Performance of different baselines under different magnitudes of confounder. The magnitude is adjusted according to probability $P$, which is the probability of a particular confounder occurring under the occurrence of specific causal factors.  The numbers in ``( )'' indicate relative performance compared to ERM: \textcolor{green}{green} for improvement, \textcolor{red}{red} for inferiority.}
	\label{tab:t2}
	\begin{center}
		\begin{tabular}{l|cccccc}
			\hline\rule{-3pt}{10pt}
			Method  &  P=5\%  & P=20\%  & P=40\%  & P=60\% & P=80\% & P=100\%  \\
			\hline\rule{-3pt}{10pt}
			\text{ERM} & 34.21$\pm$1.56  & 28.86$\pm$1.17 & 25.94$\pm$1.63 & 24.43$\pm$1.40 & 23.15$\pm$1.10 & 22.62$\pm$1.79\\
   			\text{ASAP} & 31.54$\pm$1.67 (\textcolor{red}{-2.67}) & 26.05$\pm$1.40 (\textcolor{red}{-2.81}) & 23.62$\pm$1.23 (\textcolor{red}{-2.32}) & 23.24$\pm$1.08(\textcolor{red}{-1.19}) & 22.71$\pm$1.48 (\textcolor{red}{-0.44}) & 22.35$\pm$1.04 (\textcolor{red}{-0.27})\\
      \hline
      \hline\rule{-3pt}{10pt}
			\text{DIR } & 38.54$\pm$0.99(\textcolor{green}{+4.33}) & 32.62$\pm$1.29(\textcolor{green}{+3.76}) & 27.15$\pm$1.38(\textcolor{green}{+1.21}) & 26.86$\pm$0.87(\textcolor{green}{+2.43}) & 24.68$\pm$0.94(\textcolor{green}{+1.53}) & 20.71$\pm$1.17(\textcolor{red}{-1.91})\\
			\text{CIGA} & \text{45.16$\pm$1.29}(\textcolor{green}{+10.95}) & 40.48$\pm$1.08(\textcolor{green}{+11.62}) & 26.06$\pm$0.86(\textcolor{green}{+0.12}) & 24.74$\pm$1.04(\textcolor{green}{+0.31}) & 23.05$\pm$1.28(\textcolor{red}{-0.10}) & 19.76$\pm$0.95(\textcolor{red}{-2.86}) \\
			\text{RCGRL} & 34.94$\pm$0.96(\textcolor{green}{+0.73}) & 31.96$\pm$1.17(\textcolor{green}{+3.10}) &  24.83$\pm$0.69(\textcolor{red}{-1.11}) & 23.72$\pm$0.75(\textcolor{red}{-0.71}) & 23.51$\pm$0.62(\textcolor{green}{+0.36}) & 21.26$\pm$0.53(\textcolor{red}{-1.36})\\
   			\text{DISC} & 41.25$\pm$0.83(\textcolor{green}{+7.04}) & 40.00$\pm$0.98(\textcolor{green}{+11.14}) &  37.00$\pm$0.92(\textcolor{green}{+11.06}) & 35.15$\pm$1.35(\textcolor{green}{+10.72}) & 33.50$\pm$1.08(\textcolor{green}{+10.35}) & 23.60$\pm$0.64(\textcolor{green}{+0.98})\\
			\hline 
		\end{tabular}
	\end{center}
\end{table*}

\begin{table*}[ht]\scriptsize
	\setlength{\tabcolsep}{10pt}
 	\caption{Performance of different baselines under different magnitudes of confounder, which is adjusted according to size. $Size$ indicates the extent to which the volume of confounder data exceeds that of the causal factor data.  The numbers in ``( )'' indicate relative performance compared to ERM: \textcolor{green}{green} for improvement, \textcolor{red}{red} for inferiority.}
	\label{tab:t3}
	\begin{center}
		\begin{tabular}{l|cccccc}
			\hline\rule{-3pt}{10pt}

			Method  &  Size=1  & Size=3  & Size=8  & Size=15 & Size=20 & Size=30  \\
			\hline\rule{-3pt}{10pt}
			\text{ERM} & 35.40$\pm$0.98 & 32.70$\pm$1.12 & 30.30$\pm$0.69 & 28.80$\pm$0.73 & 27.70$\pm$0.57 & 27.30$\pm$1.19\\
   			\text{ASAP} & 26.10$\pm$0.73(\textcolor{red}{-9.30})  & 25.40$\pm$1.49(\textcolor{red}{-7.30}) & 25.00$\pm$1.26(\textcolor{red}{-5.30}) & 24.80$\pm$1.17(\textcolor{red}{-4.00}) & 24.70$\pm$1.08(\textcolor{red}{-3.00}) & 24.20$\pm$0.59(\textcolor{red}{-3.10})\\
            \hline
      \hline\rule{-3pt}{10pt}
			\text{DIR} & 35.80$\pm$0.86(\textcolor{green}{+0.40}) & 33.70$\pm$1.13(\textcolor{green}{+1.00}) & 30.50$\pm$0.96(\textcolor{green}{+0.20}) & 29.40$\pm$0.73(\textcolor{green}{+0.60}) & 28.30$\pm$0.82(\textcolor{green}{+0.60}) & 25.50$\pm$0.79(\textcolor{red}{-1.80})\\
			\text{CIGA} & 28.25$\pm$1.31(\textcolor{red}{-7.15}) & 26.40$\pm$0.76(\textcolor{red}{-4.60}) & 24.90$\pm$0.94(\textcolor{red}{-5.40}) & 24.30$\pm$1.24(\textcolor{red}{-4.50}) & 24.20$\pm$0.71(\textcolor{red}{-3.50}) & 23.00$\pm$0.84(\textcolor{red}{-4.30}) \\
			\text{RCGRL} & 32.20$\pm$0.93(\textcolor{red}{-3.40}) & 28.10$\pm$0.65(\textcolor{red}{-4.60}) &  27.90$\pm$1.17 (\textcolor{red}{-2.40})& 27.80$\pm$0.71(\textcolor{red}{-1.00}) & 25.50$\pm$1.06(\textcolor{red}{-2.20}) & 24.50$\pm$1.20(\textcolor{red}{-2.80}) \\
   			\text{DISC} & 41.70$\pm$0.85(\textcolor{green}{+6.30}) & 40.10$\pm$1.06(\textcolor{green}{+7.40}) &  39.10$\pm$0.62(\textcolor{green}{+8.80}) & 38.40$\pm$1.24(\textcolor{green}{+9.60}) & 37.10$\pm$1.18(\textcolor{green}{+9.40}) & 36.30$\pm$0.95(\textcolor{green}{+9.00}) \\
			\hline 
		\end{tabular}
  
	\end{center}
\end{table*}

\begin{table*}[ht]\scriptsize
	\setlength{\tabcolsep}{10pt}
 	\caption{Performance of different baselines under different magnitudes of confounder. The magnitude is adjusted according to the complexity of the relationship between the confounder and the causal factor. The complexity level is labeled in the first row of the table.  The numbers in ``( )'' indicate relative performance compared to ERM: \textcolor{green}{green} for improvement, \textcolor{red}{red} for inferiority.}
	\label{tab:t4}
	\begin{center}
		\begin{tabular}{l|cccccc}
			\hline\rule{-3pt}{10pt}

			Method  &  Very low   & Low  & Medium  & High & Very high & Extremely high  \\
			\hline\rule{-3pt}{10pt}
			\text{ERM} & 33.10$\pm$0.78 & 32.90$\pm$1.11 & 31.60$\pm$0.89 & 31.20$\pm$0.76 & 29.50$\pm$1.13 & 27.80$\pm$0.97\\
   			\text{ASAP} & 40.50$\pm$1.22(\textcolor{green}{+7.40}) & 38.10$\pm$0.87(\textcolor{green}{+5.20}) & 37.70$\pm$0.59(\textcolor{green}{+6.10}) & 36.40$\pm$0.71(\textcolor{green}{+5.20}) & 36.00$\pm$1.04(\textcolor{green}{+6.50}) & 34.20$\pm$0.96(\textcolor{green}{+6.40})\\
      \hline
      \hline\rule{-3pt}{10pt}
			\text{DIR} & 36.00$\pm$1.12(\textcolor{green}{+2.90}) & 35.70$\pm$0.93(\textcolor{green}{+2.80}) & 34.50$\pm$0.74(\textcolor{green}{+2.90}) & 34.30$\pm$1.16(\textcolor{green}{+3.10}) & 33.10$\pm$0.83(\textcolor{green}{+3.60}) & 33.00$\pm$1.18(\textcolor{green}{+5.20})\\
			\text{CIGA} & 32.50$\pm$0.94(\textcolor{red}{-0.60}) & 31.10$\pm$1.07(\textcolor{red}{-1.80}) & 29.90$\pm$1.18(\textcolor{red}{-1.70}) & 29.80$\pm$0.86(\textcolor{red}{-1.40})  & 25.50$\pm$1.23(\textcolor{red}{-4.00}) & 25.10$\pm$0.92(\textcolor{red}{-2.70})\\
   			\text{RCGRL} & 30.10$\pm$1.14(\textcolor{red}{-2.00}) & 29.00$\pm$0.98(\textcolor{red}{-3.90}) & 27.40$\pm$1.28(\textcolor{red}{-4.20}) & 27.20$\pm$1.36(\textcolor{red}{-4.00}) & 25.80$\pm$0.76(\textcolor{red}{-3.70}) & 25.30$\pm$1.07(\textcolor{red}{-2.50})\\
         	\text{DISC} & 43.65$\pm$0.96(\textcolor{green}{+10.55}) & 41.00$\pm$1.03(\textcolor{green}{+8.10}) & 39.55$\pm$1.42(\textcolor{green}{+7.95}) & 39.10$\pm$0.75(\textcolor{green}{+7.90}) & 38.95$\pm$0.83(\textcolor{green}{+9.45}) & 37.25$\pm$1.25(\textcolor{green}{+9.45})\\
			\hline 
		\end{tabular}
  
	\end{center}
\end{table*}

\begin{table*}[ht]\scriptsize
	\setlength{\tabcolsep}{10pt}
 	\caption{Performance in different datasets, including classification accuracy in Graph-SST5(ID) and Graph-Twitter, and Unbiased and Biased Spurious-Motif,and our dataset CRCG. The records with improvements compared to the original methods are highlighted in \textbf{bold}.}
	\label{tab:pf}
	\begin{center}
		\begin{tabular}{l|cccccc}
			\hline\rule{-2pt}{10pt}
						
			Method &  Graph-SST5  & Graph-Twitter & Spurious-Motif &CRCG (P=20\%) & CRCG (P=40\%) & CRCG (P=80\%)
            \\
			\hline\rule{-2pt}{10pt}
			\text{ERM} & 42.30$\pm$0.87 & 61.20$\pm$1.05 & 33.20$\pm$0.95 & 28.80$\pm$0.75 & 25.94$\pm$1.63	& 24.43$\pm$1.40  \\
			\textbf{ERM + R-CAM} & \bf{43.40$\pm$0.68} & \bf{63.70$\pm$1.21}  & \bf{35.60$\pm$1.05} & \bf{31.60$\pm$0.93} & \textbf{25.95$\pm$1.20} & 24.93$\pm$0.23 \\

            \hline\rule{-2pt}{10pt}
			\text{ASAP } & 44.50$\pm$1.34 & 61.50$\pm$0.97  & 34.90$\pm$1.25 & 26.05$\pm$1.26  & 23.62$\pm$1.23  &  22.71$\pm$1.48   \\
			\textbf{ASAP + R-CAM} & \bf{46.00$\pm$0.97} & \bf{64.10$\pm$0.52}  & 34.20$\pm$1.37 & \bf{30.80$\pm$1.32} & \textbf{28.25$\pm$0.80} & \textbf{23.10$\pm$0.40} \\
   
            \hline\rule{-2pt}{10pt}
			\text{DIR } & 44.20$\pm$1.26 & 62.80$\pm$0.97 & 43.60$\pm$0.73 & 23.60$\pm$0.84  & 27.15$\pm$0.86 & 24.68$\pm$0.94   \\
			\textbf{DIR + R-CAM} & \bf{46.00$\pm$1.60} & 62.40$\pm$0.52 & \bf{47.30$\pm$1.47} & \bf{31.30$\pm$1.47} & \textbf{30.10$\pm$2.55} & \textbf{27.68$\pm$0.48} \\

            \hline\rule{-2pt}{10pt}
			\text{CIGA}  &  44.20$\pm$1.03 &  58.90$\pm$0.77  & 34.40$\pm$0.79 & 27.40$\pm$0.98  & 26.06$\pm$0.86 & 23.05$\pm$1.28  \\ 
			\textbf{CIGA + R-CAM} &  \bf{45.40$\pm$0.82} &  \bf{60.70$\pm$1.31}  & \bf{36.00$\pm$1.03} & 27.00$\pm$0.89 & \textbf{36.03$\pm$1.13} & \textbf{25.93$\pm$0.28} \\ 

            \hline\rule{-2pt}{10pt}
			\text{RCGRL}   &  44.50$\pm$1.46 &  60.10$\pm$0.74 & 45.70$\pm$0.98 & 33.30$\pm$0.93  & 24.83$\pm$0.69 & 23.51$\pm$0.62  \\ 
   			\textbf{RCGRL + R-CAM}   &  \bf{46.50$\pm$1.08} &  \bf{63.40$\pm$0.96}  & \bf{48.50$\pm$0.75}  & \bf{34.40$\pm$1.39} & \textbf{28.75$\pm$0.65} & \textbf{24.55$\pm$0.20} \\ 
      
            \hline\rule{-2pt}{10pt}
			\text{DISC}   &  34.40$\pm$1.28 &  62.50$\pm$1.54 &    42.85$\pm$1.23 & 40.10$\pm$1.36  & 37.00$\pm$0.92 & 33.50$\pm$1.08  \\ 
   			\textbf{DISC + R-CAM}   &  \bf{38.15$\pm$1.21} &  61.70$\pm$0.89 &    \bf{47.52$\pm$0.97}  & \bf{41.05$\pm$1.05} & \textbf{39.18$\pm$1.18} & \textbf{34.40$\pm$3.10} \\ 
			\hline 
		\end{tabular}
    \vskip -0.1in
	\end{center}

\end{table*}



\section{Methodology}
Drawing from theoretical analysis and experimental outcomes, we propose to emphasize the model's causal model capability by directly applying influence to the model's outputs. Additionally, such influence should be applied through a probabilistic perspective. Specifically, in light of Corollary \ref{cly:m}, the actions we apply should aim to maximize the independence between $C$ and $f(G)$. Subsequently, we must introduce certain priors to guide our operations. Building upon Theorem \ref{lma:da}, we can derive the following corollary: 
\begin{corollary}
Under the conditions specified in Theorem \ref{lma:da}, the following inequality holds:
\begin{gather}
    I(C;Y) \leq I(X;Y)
\end{gather}
where $C$ denotes the set of confounders within $G$, $Y$ is the ground-truth label, and $X$ denotes the set of causal factors within $G$.
\label{cly:m}
\end{corollary} 
The proof can be found in \textbf{Appendix} A.5. Therefore, We can draw the conclusion that features with lower mutual information with the ground-truth labels tend to possess a higher propensity of being confounders. Conversely, features that possess higher mutual information with ground-truth labels are inclined to exhibit a diminished likelihood of being confounders. However, formal computation of the aforementioned mutual information is challenging, we need an alternative solution. As the mutual information between two variables indicates the extent to which observing one variable reduces the uncertainty about the other variable. Therefore, we treat the features that appear consistently in graph samples of the same category as causal factors. Furthermore, we treat the features that appear in graph samples of different categories as confounders. Next, we proceed to illustrate how we leverage this conclusion to conduct causal optimization of the model.

Specifically, for the graph training dataset $\{G_{i}\}_{i=1}^n$, we can acquire the node representations with GNN $g_{\phi}(\cdot)$. Formally, we have:
\begin{gather}
    \bm{Z}_{i} = g_{\phi}(G_{i}),
\end{gather}
where $\bm{Z}_{i}$ denotes the set of output node representations of graph $G_{i}$, $\bm{Z}_{i,j}$ denotes the node representation of node $j$ within $G_{i}$. We employ the function $r_{\psi}(\cdot)$ to perform pooling on node representations, followed by generating predictions and computing the cross-entropy-based loss. The loss function can be formulated as follows:
\begin{gather}
	 \mathcal{L}_{CE} =  \sum_{i=1}^{n} \mathcal{H}\Big(r_{\psi}(\bm{Z}_{i}),Y_{i}\Big),
\end{gather}

Where $\mathcal{H}(\cdot)$ calculates the cross entropy loss. Subsequently, we partition the node representations $\{\bm{Z}_{i}\}^{n}_{i=1}$ based on their respective class labels and the correctness of classification results. For ease of comprehension, we use $c$ to denote the class, $c \in \{1,2,...,m\}$, $m$ is the number of classes. For samples with ground-truth labeled class $c$, we select all those graph samples that are correctly classified as class $c$ and construct a matrix $S^{+}_{c}$ with their corresponding node representations. $S^{+}_{c} \in \mathbb{R}^{v \times h}$, $v$ denotes the number of node representations that used to build $S^{+}_{c}$, $h$ denote the length of representation vectors. Likewise, we identify all incorrectly classified samples and assemble their node representations into a matrix $S^{-}_{c}$. Then, we calculate matrix $S_{c}^{M}$ with the following equation:
\begin{gather}
    S_{c}^{M} = \Big(S^{+}_{c} \cdot (S^{-}_{c})^{T}\Big) \odot \Big(u(S^{+}_{c}) \cdot (u(S^{-}_{c}))^{T}\Big),
    \label{eq:sim}
\end{gather}
where $\odot$ denotes the Hadamard Product. $u(\cdot)$ can be formulated as follows:
\begin{gather}
    u(S) = \begin{bmatrix} \ \frac{1}{|\bm{s}_{1}|} & ... & \frac{1}{|\bm{s}_{n}|} \ \ \end{bmatrix} ^{T},
\end{gather}
where $S$ is a matrix, and $\bm{s}$ denotes the row vectors. Equation \ref{eq:sim} allows for the computation of the cosine similarity between all the representation vectors in $S^{+}_{c}$ and $S^{-}_{c}$, where $S_{c}^{M}$ represents the resulting similarity matrix.

Then, we select the elements within $S_{c}^{M}$ that are larger than a hyperparameter $\tau$, and mark them as ``anchor node representations''. We traverse through all the similarity matrices corresponding to different categories to label all the anchor node representations. The anchor node representations represent the features that consistently appear in graph samples of the same category. As discussed before, we consider these features to be more reliable and less susceptible to confounders compared to other features. For each graph sample $G_{i}$, we denote the set of its anchor node representations as $\bm{X}_{i}$. Subsequently, we compute the feature emphasis loss $\mathcal{L}_{a}$ based on $\bm{X}_{i}$.
\begin{gather}
	 \mathcal{L}_{a} =  -\sum_{i=1}^{n} s\Big(\text{pool}(\dot{\bm{X}_{i}}),\text{pool}(\bm{Z}_{i})\Big),
\end{gather}
where $s(\cdot)$ is the function that calculates the cosine similarity between variables. $\text{pool}(\cdot)$ denotes the function that conduct pooling operation. $\dot{\bm{X}_{i}}$ denotes that $\bm{X}_{i}$ is detached from back propagation. Therefore, $\mathcal{L}_{a}$ encourages other node representations to become more similar to the anchor node representations, thereby emphasizing the correct and persistent features across the graph samples.

Next, we design the model to ignore information that may be affected by confounders. For all samples classified as class $c$, we extract the node representations of those correctly classified graph samples and assemble them into a matrix $I^{+}_{c}$. Like $S^{+}_{c}$, $I^{+}_{c} \in \mathbb{R}^{l \times h}$, $l$ denotes the number of node representations that are used to construct $I^{+}_{c}$, $h$ denote the length of representation vectors. Then, we construct matrix $I^{-}_{c}$ from the node representations of those misclassified samples. We calculate matrix $I_{c}^{M}$ with the following equation:
\begin{gather}
    I_{c}^{M} = \Big(I^{+}_{c} \cdot (I^{-}_{c})^{T}\Big) \odot \Big(u(I^{+}_{c}) \cdot (u(I^{-}_{c}))^{T}\Big).
\end{gather}
Then, we select the elements within $I_{c}^{M}$ that are larger
than $\tau$, and mark them as ``deceptive node representations''. Like the above, We traverse through all the similarity matrices corresponding to different categories to label all the deceptive node representations. The deceptive node representations contained in $I_{c}^{M}$ represent the representations that appear in graph samples classified by the model as the category $c$. Additionally, these representations also appear in the node representations of samples misclassified as class $c$, suggesting that they may be influenced by confounders that are probabilistically correlated with the labels under some scenarios. Therefore, we aim for the model to disregard these representations. For each graph sample $G_{i}$, we denote the set of its anchor node representations as $\bm{C}_{i}$. Subsequently, we compute the feature ignoring loss $\mathcal{L}_{i}$ based on $\bm{C}_{i}$.
\begin{gather}
	 \mathcal{L}_{i} =  \sum_{i=1}^{n} s\Big(\text{pool}(\dot{\bm{C}_{i}}),\text{pool}(\bm{Z}_{i})\Big),
\end{gather}  
$\mathcal{L}_{i}$ encourages other node representations to become less similar to the deceptive node representations, thereby disregarding these representations.

We sum up $\mathcal{L}_{i}$ and $\mathcal{L}_{a}$ as the causal enhance loss:
\begin{gather}
	 \mathcal{L}_{c} =  \mathcal{L}_{i} + \mathcal{L}_{a}.
\end{gather}  
$\mathcal{L}_{c}$ can be incorporated into the training of any GNN model to enhance its causality. The overall training loss is a summation of our proposed loss $\mathcal{L}_{c}$ and the original model loss. Furthermore, we can adjust the hyperparameter $\tau$ to control the extent of the module's influence, thus adapting to different datasets. Our proposed R-CAM is only adopted for training and removed for testing.

\section{Experiments}
\begin{figure}[ht]
	\centering
	\subfigure[ERM]{
		\begin{minipage}{0.21\textwidth} 
			\includegraphics[width=\textwidth]{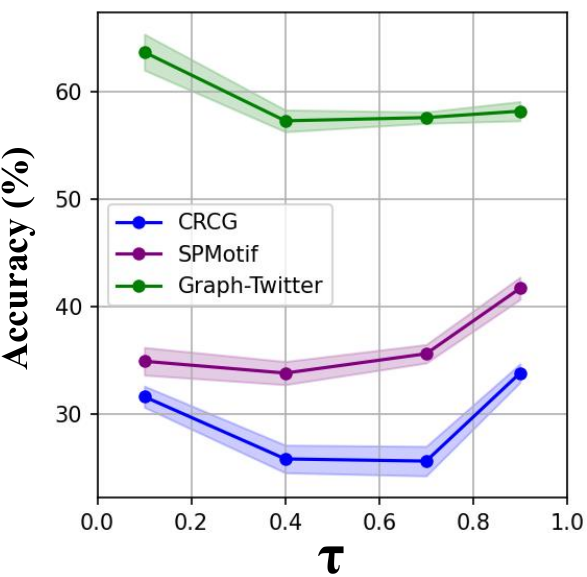} 
		\end{minipage}
	}
	\subfigure[DIR]{
		\begin{minipage}{0.21\textwidth}
			\includegraphics[width=\textwidth]{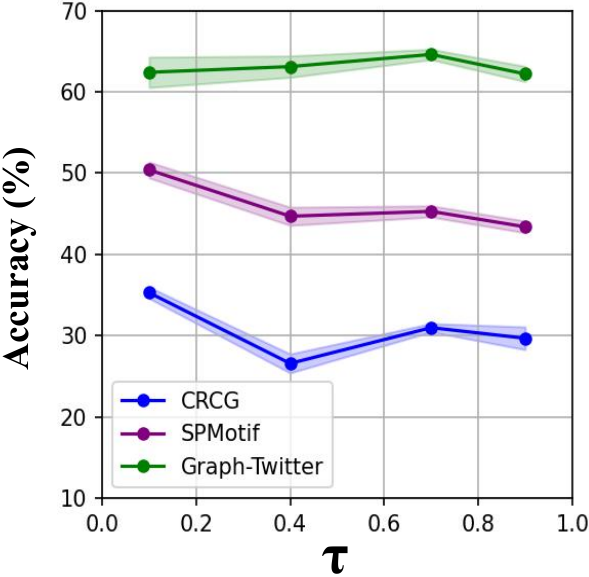} 
		\end{minipage}
    }	
	\caption{ Performance of ERM and DIR across datasets with varying hyperparameters $\tau$, where the semi-transparent part indicates the standard deviation.}
	\label{fig:charts}
\end{figure}

\begin{table}[ht]\small
    \centering
    \caption{ The averaged Friedman test results, encompass experiments both with and without confounders, and comparative experiments with state-of-the-art methods.}
    \begin{tabular}{l|ccc}
    \hline\rule{-2pt}{10pt}
        \multirow{2}*{Experiment} & Friedman & \multirow{2}*{P value} & Significant  \\
         & Statistic &  & Differences \\
        \hline
        \hline\rule{-2pt}{10pt}
         No confounder & 24.54 & 1.71 $\times$ $10^{-4}$  & exist\\
         With confounder & 23.17 & 3.13 $\times$ $10^{-4}$ & exist\\
         Comparison & 45.00 & 9.22 $\times$ $10^{-7}$ & exist\\
         \hline
    \end{tabular}
    \label{tab:ft}
\end{table}

\begin{table}[ht]\small
    \centering
    \caption{CPU time overhead for different methods, measured in seconds.}
    \begin{tabular}{l|ccc}
    \hline\rule{-2pt}{10pt}
        Methods & Graph-SST5 & Graph-Twitter & CRCG \\
        \hline
        \hline\rule{-2pt}{10pt}
        ERM & 101.22 & 22.54 & 25.00\\
        ERM+R-CAM &	103.32 & 23.98 & 27.20\\
         DIR & 194.95 & 93.83  & 91.32\\
         DIR+R-CAM & 204.23 & 95.77 & 94.49\\
         CIGA & 25.60 & 4.44 & 4.85 \\
         CIGA+R-CAM & 26.39 & 4.61 & 4.98 \\
         \hline
    \end{tabular}

    \label{tab:cpu}
\end{table}

\subsection{Effect analysis}
\subsubsection{Settings}
We evaluated our method on various datasets including: 1) Graph-SST5 \cite{dataset:sst}, 2) Graph-Twitter \cite{dataset:sst}, and 3) Spurious-Motif \cite{DBLP:conf/iclr/WuWZ0C22} under different bias, 4) our proposed CRCG. Further details are in \textbf{Appendix} D.3. We integrated the R-CAM method into different baselines to conduct before-and-after comparative experiments. Further details are in \textbf{Appendix} D.1 and D.2. 

\subsubsection{Results}
Results are summarized in Table 6. After integrating R-CAM, the majority of algorithms showed varying degrees of accuracy improvement across datasets. This validates the effectiveness of R-CAM in emphasizing causal information within the data.



\subsection{Statistically Significance Analysis}
To demonstrate the statistical significance of our experiments, we conducted the Friedman test on model performance experiments. The results are demonstrated in Table \ref{tab:ft}. We can observe that according to the results, the statistically significant differences generally exist with a significance level of 0.01. Furthermore, based on the results obtained from averaging the outcomes of five experimental runs, it is clear that our method outperforms the baseline methods to a significant extent. To illustrate, when compared to its own baseline, ERM, DIR shows an average accuracy improvement of 5.1\%. However, with the addition of R-CAM, the accuracy improvement increases to 15.1\%. This underscores the statistical significance of R-CAM's effectiveness.

\subsection{Computation Cost}

To analyze the computational cost of R-CAM, we measure the CPU time dedicated to computation. As indicated in the experimental results presented in Table \ref{tab:cpu}, the integration of R-CAM into ERM results in an average increase of 5.75\% in CPU time overhead. Similarly, DIR and CIGA exhibit average increases of 3.8\% and 3.1\% in CPU time overhead, respectively. From these findings, we infer that the CPU time overhead associated with our proposed method is relatively modest.

\subsection{Evaluation on Module Structure}
We further evaluated R-CAM by adjusting the similarity threshold $\tau$ for the ERM and DIR \cite{DBLP:conf/iclr/WuWZ0C22} algorithms on the CRCG, Spurious-Motif, and Graph-Twitter datasets. Figure \ref{fig:charts} shows that the highest accuracy varies across datasets with different thresholds. This demonstrates that by adjusting the hyperparameter $\tau$, our model can adapt to various datasets. Simultaneously, this phenomenon also substantiates that the distribution of confounders in different datasets is indeed characterized by substantial disparities.

\section{Conclusion}
This paper introduces an innovative synthetic dataset, CRCG, designed specifically for evaluating the causal modeling capabilities of GNNs. Subsequent to the dataset introduction, we conduct thorough theoretical and experimental analyses, culminating in the introduction of a lightweight GNN causal enhancement module known as R-CAM. The efficacy of R-CAM is validated through a series of comprehensive experiments.

\section{Acknowledgement}
The authors would like to thank the editors and reviewers for their valuable comments. This work is supported by the CAS Project for Young Scientists in Basic Research, Grant No. YSBR-040, and the Youth Innovation Promotion Association CAS.

\bibliography{aaai24}

\clearpage
\appendix
\section{A. Proofs.}
\subsection{A.1. Proof of Theorem \ref{lma:da}}


We demonstrate the theorem with the Causal Markov Condition \cite{pearl2000models}. According to the definition of the confounder, we have $C \not \! \perp \!\!\!\perp Y$, furthermore, $C$ has no causal relationship with $Y$. If the graph generation can be treated as a Markov process, then the causal model including $G$, the parents of $G$, and $Y$ can be viewed as a Markovian causal model.

Then, according to Causal Markov Condition, in a Markovian causal model, each variable is independent of all its nondescendants, given its parents. Based on the discussion above, we can view our problem as a Markovian causal model as well. As $C$ has no causal relationship with $Y$, therefore $C$ contains no ancestor of $Y$. Then if given the parents of $C$, $C \perp \!\!\!\perp Y$. Furthermore, as $C$ contains none descendant of $X$ or the ancestors of $X$, given the parents of $C$ or not does not influence the dependency between $C$ and $Y$, therefore $C \perp \!\!\!\perp Y$ under all conditions, which is in contrary with $C \not \! \perp \!\!\!\perp Y$. Therefore, $C$ contains descendants of $X$ or the ancestors of $X$, the theorem is proved.


\subsection{A.2. Proof of Theorem \ref{thm:noc}}
According to the theorem, $G$ only consists of two parts: $X$ and $Z$. $X$ denotes the causal factors that determined the value of $Y$. In other words, in a causal model which models the current scenario, there exists a directed edge from $X$ to $Y$. $Z$ denotes random noise that $Z \perp \!\!\!\perp Y$. 

As the conditional independence $(S \perp \!\!\!\perp Y | G)$ holds, there are no back-door paths from $Y$ to $G$, and thus represent there is no confounder between $X$ and $Y$. According to the property of causal effect \cite{pearl2000models}, we have:
\begin{gather}
    P(Y|\hat{X}) = P(Y|X).
\label{eq:yx}
\end{gather}
As we have $Z \perp \!\!\!\perp Y$, therefore:
\begin{gather}
    P(Y|\hat{G}) = P(Y|\hat{X}) = P(Y|X) = \nonumber\\ P(Y|X,Z) = P(Y|G).
\end{gather}
Hence, if a set of parameters $\theta^{*}$ exists, such that the GNN $f_{\theta^{*}}(\cdot)$ is capable of capturing the underlying relationship between the variables $G$ and $Y$, then $\sum_{i=1}^{n} \mathcal{H}\big(f_{\theta}(G_{i}),Y_{i}\big)$ attains a minimum value if:
\begin{gather}
    P(Y|f_{\theta}(G)) = P(Y|G). 
\end{gather}
i.e.:
\begin{gather}
    P(Y|f_{\theta}(G)) = P(Y|\hat{G}). 
\end{gather}
According to definition \ref{def:ce}, $P(Y|\hat{G})$ is causally estimable, and the theorem is proved.

\subsection{A.3. Proof of Theorem \ref{thm:do}} 
As $C$ has no causal relationship with $Y$, therefore $C$ doesn't open another causal root towards $Y$. Then Equation \ref{eq:yx} within the proof of Theorem \ref{thm:noc} still holds. We have:
\begin{gather} 
    P(Y|X) = P(Y|\hat{X}). 
\end{gather}
We also have:
\begin{gather} 
    P(Y|\hat{X}) \neq P(Y|\hat{X},C). 
\end{gather}
If there exist confounder $C$ in $G$, we have $C \not \! \perp \!\!\!\perp  Y$ and $ C \neq X $. Furthermore, we have:
\begin{gather} 
    P(Y|X,C) \neq P(Y|X) = P(Y|\hat{X}) = \nonumber\\ P(Y|\hat{X},C) = P(Y|\hat{G}). 
\end{gather}
When $\sum_{i=1}^{n} H\big(f_{\theta}(G_{i}),Y_{i}\big)$ reaches minimal, we have: 
\begin{gather} 
    P(Y|f_{\theta}(G)) = P(Y|G) = P(Y|X, C) \neq P(Y|\hat{G}) 
\end{gather} 
According to Definition \ref{def:ce}, $P(Y|\hat{G})$ is not causally estimable. The first conclusion of the theorem has been proven. 

As for the second conclusion of the theorem, we have: 
\begin{gather} 
    P(Y|\hat{X}) = P(Y|X), 
\end{gather} 
where $X = G \setminus C$, i.e. the data within $G$ that denotes the causal factors. Then, if intervention on $C$ is possible, we have: 
\begin{gather} 
    P(Y|\hat{X}, do(C=C_{i})) = P(Y|X,do(C=C_{i})). 
\end{gather} 
Furthermore, the following equation holds: 
\begin{gather}
	 \theta^{*} = \arg\min_{\theta}\Big( \sum_{i=1}^{n} \mathcal{H}\big(f_{\theta} (\textbf{X}_{i}\cap do(C=C_{i})),Y_{i}\big)\Big),
\end{gather}
where $G=X\cap C$. In a causal structure, whether $C$ contains descendants of $X$ or the ancestors of $X$, intervention $do(C=C_{i})$ could then ensure a cut-off of such relationships between them. Therefore, for certain graph data $G_{i}$ we have:
\begin{gather} 
    P\big(Y|f_{\theta^{*}}(G_{i})\big)=
    P\big(Y|f_{\theta^{*}}(\textbf{X}_{i}\cap do(C=C_{i}))\big) \nonumber\\ = P(Y|\hat{X}_{i}, do(C=C_{i})) = P(Y|\hat{X}_{i}).
\end{gather} 
As $X$ denotes the causal factors within $G$ and $C$ denotes the confounders, $ do(C)=\widetilde{C}, \forall \widetilde{C} \in \mathcal{C}$ is feasible, $f_{\theta^{*}}(\cdot)$ could measure the causal relationship between $G$ and $Y$ while ignoring the confounders given any $G$, the theorem is proved.


\subsection{A.4. Proof of Corollary \ref{cly:cd}}
As we have $f(G) \perp \!\!\! \perp C \ | \ T$, therefore we can conclude that given $T$, $f(G)$ won't altered according to $C$, i.e., $f(G)$ will output the same value if given $X$ and $Z$ regardless of the value of $C$.

Therefore, if given $T$, and based on Equation \ref{eq:lr}, the following equation holds:
\begin{gather}
   \sum_{i=1}^{n} \mathcal{H}\big(f_{\theta}(G_{i}),Y^{*}_{i}\big) \perp \!\!\! \perp C.
\end{gather}

Therefore, the value of $\theta^{*}$ in Equation \ref{eq:lr} will not be influenced by $C$, as well as $Z$. Furthermore, we have:
\begin{gather}
    I\big((f(G);X \ | \ T\big) = I\big(f(G);X\big),
\end{gather}
i.e. $T$ does not influence $f(\cdot)$ in modeling the static relationship between $X$ and $Y$. Then, if there exist $\theta^{*}$ that enable the follows:
\begin{gather}
f_{\theta^{*}}(G)=P(Y|\hat{G})=P(Y|\hat{X})=P(Y|X)
\end{gather}
, $\theta^{*}$ can be calculate through Equation \ref{eq:lr} given $T$. Because only $X$ will influence the value of $\theta$ and $T$ does not influence $f(\cdot)$ in modeling the static relationship between $X$ and $Y$. The corollary is proofed.

\subsection{A.5. Proof of Corollary \ref{cly:m}}
According to Theorem \ref{lma:da}, the set of confounders $C$ in $G$ must be descendants of the set of causal factors $X$ in $G$ or their ancestors. Therefore, we have:
\begin{gather}
    I(X;C) \leq H(X).
\end{gather}
As $X$ consist of all the causal factors within $G$ that holds causal relationship with $Y$, we have:
\begin{gather}
    I(X;Y) = H(Y) = H(X).
\end{gather}
Furthermore, we have the maximize value of $I(C;Y)$ is $H(Y)$, and:
\begin{gather}
    I(C;Y|X) = 0.
\end{gather}
Therefore, the following inequality holds:
\begin{gather}
    I(C;Y) \leq I(X;Y).
\end{gather}
And the equality only holds when the distributions of $C$ and $X$ are exactly the same.


\section{B. Details of CRCG Dataset}
\subsection{Overview}
The CRCG dataset we introduce represents a novel endeavor in generating data that encapsulates causal relationships reflective of logical truths adhered to by the dataset. Formulated based on the generative approach elucidated in DIR, CRCG encompasses two distinct categories: graph classification and node classification synthetic datasets. The former comprises motifs, or subgraphs possessing specific properties, along with inter-motif relations. This formulation aims to enhance the evaluation of model causality within whole-graph classification tasks. Meanwhile, the latter category encompasses node classification datasets that facilitate the creation of highly feature-rich nodes adhering to adjustable and definable patterns. These nodes are linked by relations designed in accordance with predefined relationship types, thereby culminating in the construction of the complete graph.
\subsection{Dataset specific composition}
For the graph classification datasets, we devised 25 motif types based on the dataset generation approach outlined in DIR. The shape of each motif is adjustable through parameters, and node features can be set following specific patterns. For instance, fixed means and standard deviations can be established. Subsequent to motif generation, we designed 4 connection relationships to facilitate the overall graph construction. The types of motifs and relationships are explicitly detailed as follows:
\subsubsection{Motif generation method}
\begin{itemize}
\item The 10 types of motifs with branches can generate the following types: star-shaped, path-shaped, fan-shaped, acute polygon, random bipartite graph, tree-shaped, trident-shaped, cone-connected graph, chain with bypass, and partial polygon. For all motif types, the number of nodes, branches, and node features can be adjusted and customized.
\item The 15 types of motifs without branches can generate the following types: complete graph, cycle graph, double cycle graph, triangle graph, ring-shaped graph, diamond graph, H-shaped graph, wheel graph, hourglass graph, DCD trident-shaped graph, circular cross graph, ladder graph, star graph, triangle graph, and cross-arm graph. For all motif types, the number of nodes and node features can be adjusted and customized.
\end{itemize}
\subsubsection{Motif connection method}
\begin{itemize}
\item Adjacent: Signifying that two motifs are connected together by an edge.
\item Cross: Indicating that two motifs share some vertices, and the number of shared vertices can be configured.
\item Entangled: Referring to two motifs connected together by multiple edges, and the number of edges can be configured.
\item Containment: Denoting that one of the motifs, with fewer vertices, is completely contained within the other motif.
\end{itemize}
For the node classification datasets, leveraging the dataset generation approach provided by DIR, we formulated 25 distinct node generation methods. After specifying the creation of a designated quantity of certain node types, we subsequently devised two criteria for establishing edges. Then, we selected one of these criteria to establish the edge relationships among these nodes, thereby achieving the comprehensive construction of the entire graph. The types of node generation methods and edge establishment criteria are explicitly elucidated as follows:
\subsubsection{Node generation method}
\begin{itemize}
\item 15 methods generated based on mean and standard deviation: node features conform to normal distribution, uniform distribution, exponential distribution, log-normal distribution, gamma distribution, beta distribution, Weibull distribution, Laplace distribution, logistic distribution, Rayleigh distribution, Pareto distribution, Cauchy distribution, negative binomial distribution, Gumbel distribution, and Gompertz distribution. For all methods, the number of nodes, as well as the mean and standard deviation of node features, can be adjusted.
\item 10 methods generated based on sequences: node features follow arithmetic sequence, geometric sequence, Fibonacci sequence, square number sequence, cubic number sequence, prime number sequence, triangular number sequence, rectangular number sequence, binomial coefficient sequence, and Hamiltonian sequence. For all methods, the number of nodes, the dimension of node features, and the parameters required for sequence computation can be adjusted.
\end{itemize}
\subsubsection{Edge Creation Criteria}
\begin{itemize}
\item Similar edges: Nodes determine their similarity autonomously (using cosine similarity), and if the similarity surpasses a fixed threshold, an edge is formed.
\item Partially similar edges: Nodes determine the similarity of specific dimensions (using cosine similarity), and if the similarity surpasses a fixed threshold, an edge is formed.
\end{itemize}

\subsection{Dataset generation method}
We leverage existing resources to construct experimental datasets through dataset composition. Each such construction is referred to as a dataset composition instance. Since our subgraphs and their interconnections can be customized, theoretically, numerous dataset composition instances can be generated, each constituting a distinct dataset configuration. The methodology for constructing dataset composition instances is detailed as follows: a list is maintained, outlining methods for graph and node generation, as well as motif linking. This list is treated as input. When generating the i-th graph sample, a random integer $m$ is selected from the range of 1 to 5. Then, the $m$-th motif generation method is chosen from methods 1 to 5. Subsequently, another motif generation method is selected from methods 6 to 10. Method $m+5$ is chosen with an 80\% probability, while the others are chosen with a 20\% probability. The index of this selected method is denoted as $k$. The motif generation methods $m$ and $k$ are concatenated using a randomly selected relation. This process completes the generation of the foundational graph structure. Following this, motif generation method $m$ is parameterized by a, and motif generation method $k$ is parameterized by $f(a)$. Similar to the aforementioned connection process, two additional motifs are added, thus completing the construction of the entire graph. Finally, by setting the label of the $i$-th graph sample as $h(m,a)$, corresponding labels (category numbers) are generated for the graph data.Meanwhile, the data set additionally includes the function of adding random noise data as follows:
\begin{itemize}
\item Randomly delete or create a fixed number of edges. 
\item Randomly delete a certain number of nodes.
\item Randomly create a certain number of nodes and connect these nodes randomly to the existing graph.
\end{itemize}

\section{C. Settings and Details of Evaluation Experiments}
\subsection{C.1. Baselines}
\subsubsection{ERM}
The core idea of the ERM algorithm is to optimize model parameters by minimizing the empirical risk, enabling the model to better capture the relationships between nodes and the structure of the graph data. This involves utilizing an appropriate loss function to quantify the disparity between the model's predictions and the actual labels and optimizing the model on the training data to reduce this loss.
\subsubsection{ASAP}
ASAP\cite{ranjan2020asap} is an adaptive structure-aware pooling technique designed for learning hierarchical graph representations. This algorithm aims to effectively capture the hierarchical features of nodes by performing adaptive node importance pooling within the graph. It combines node features, connectivity information, and graph structure to compute importance scores for each node. Based on these scores, it selectively pools nodes, preserving crucial information and context.
\subsubsection{DIR}
DIR\cite{DBLP:conf/iclr/WuWZ0C22} solves the problem that traditional rationalization models often rely on data bias and shortcut feature explanation predictions, ignoring key causal patterns. DIR introduces multiple intervention distributions, intervenes in the training distribution, approaches the causal principle that remains unchanged in different distributions, eliminates unstable false patterns, and constructs an essentially explainable GNN model.
\subsubsection{CIGA}
CIGA\cite{DBLP:conf/nips/0002ZB00XL0C22} achieves OOD generalization under various distribution changes by capturing the intrinsic invariance of graph data. The framework uses a causal model to describe potential distribution changes, emphasizing the subgraph that contains the largest information related to the cause of the label, and proposes the goal of information theory to extract the expected subgraph that maintains the invariant information within the class to the greatest extent, so as to realize the invariance of distribution changes. Responsive representation learning.
\subsubsection{RCGRL}
RCGRL\cite{DBLP:conf/aaai/GaoLQSXZ023} aims to learn a robust graph representation against confounding effects, so as to avoid the interference of semantic information in the graph to the model learning. By introducing an active method to generate instrumental variables under unconditional moment constraints, the graph representation learning model can eliminate confounding factors, thereby obtaining discriminative information that is causally associated with downstream predictions.
\subsubsection{DISC}
DISC\cite{fan2022debiasing} solves the problem of the limited generalization ability of graph neural network on biased data. The framework consists of utilizing an edge mask generator to split the input graph into causal and biased subgraphs. The two GNN modules encode relevant information respectively, and integrate causal and bias-aware loss functions to further reduce the correlation between causal variables and bias variables, thereby improving the generalization ability of the model.

\subsection{C.2. Experimental Settings}
\subsubsection{Test with no confounders}
To assess the impact of GNN's ability to distinguish underlying causes in the presence of unrelated information interference, we conducted experiments across three scenarios. These scenarios encompass graph data with only genuine causal relationships to the labels, the introduction of one set of unrelated graph data noise, and the introduction of two sets of unrelated graph data noise.
\subsubsection{Test with varying probabilities of causal relationships}
To examine the effect of significant occurrences of spurious causal relationships at varying probability levels on GNN's capability to discern genuine causal factors, we conducted experiments across six scenarios. These scenarios involve graph data with false causal relationships to the labels, appearing in the training set at probabilities of 5\%, 20\%, 40\%, 60\%, 80\%, and 100\%. The sizes of these instances were matched as closely as possible to those with authentic causal relationships to the labels. In the test set, they appeared randomly similar to graph data unrelated to the labels.
\subsubsection{Test with  changing size of confounders}
To investigate the impact of significant occurrences of spurious causal relationship instances at varying data volumes on GNN's ability to distinguish genuine causal factors, we conducted experiments across six scenarios. In these scenarios, instances of "graph data with false causal relationships to the labels" were introduced into the training set with a fixed probability of 50\%, while in other cases, they appeared randomly similar to graph data unrelated to the labels. The sizes of these instances were set to be 1/3/8/15/20/30 times that of instances with "graph data with authentic causal relationships to the labels." In the test set, they appeared randomly similar to graph data unrelated to the labels.
\subsubsection{Test with changing complexity relation}
To assess the impact of varying levels of prominence of spurious causal relationships on GNN's ability to discern authentic causal factors, we conducted experiments across six scenarios. In these scenarios, instances of "graph data with false causal relationships to the labels" were introduced into the training set with differing degrees of prominence categorized as Extremely High, Very High, High, Medium, Low, and Very Low. The sizes of these instances were equivalent to those with "graph data with genuine causal relationships to the labels." In the test set, they appeared randomly similar to graph data unrelated to the labels.

\subsection{C.3. Dataset Construction Details}
To meet the above four experimental condition settings, we set the different types of data construction rules in the graph dataset as:
\begin{itemize}
\item Graphs with Genuine Causal Relationships to Labels: Selection is made from motifs 1 through 5, with labels denoted as Y in the range [0, 1, 2, 3, 4], encompassing five categories. For instance, Y=0 corresponds to the interconnection of motif1 and motif2; Y=1 corresponds to the connection of motif1 with motif3, followed by the crossing of motif5; Y=2 corresponds to the entanglement of motif1 with motif2, crossed by motif5; Y=3 corresponds to motif5; and Y=4 corresponds to the crossing of motif3 and motif4. All parameters are adjustable within a limited range.
\item Graphs with Spurious Causal Relationships to Labels: Selection is made from motifs 6 through 10, and these motifs are associated with motifs 1 through 5 with a certain probability. In other words, if motifs 1, 2, 3, 4, or 5 are present, then motifs 6, 7, 8, 9, or 10 can appear based on the experimentally set probability. For instance, if motif1 is associated with motif6, they form a one-to-one correspondence.
\item Graphs Irrelevant to Labels: A motif is randomly selected from the remaining motifs, and it is connected in a manner similar to the previous motifs, but with a larger size (number of nodes). Additionally, 10\% edge noise and 10\% node noise interference are introduced during graph construction.
\end{itemize}
After building the above data set, adjust the data for different experimental scenarios to complete the test. The key adjustment details are as follows:
\begin{itemize}
\item In the "no confounder" scenario, the specific procedure for introducing noise involves connecting motifs unrelated to the labels to the graph data that has been generated with genuine causal relationships. The number of noise instances to be added determines the corresponding quantity of motifs to be connected.
\item In the context of probability-based testing scenarios, we selected five distinct types of motifs, associating them with graph data possessing genuine causal relationships through predefined probabilities. When a specific motif type (such as motif1) is present within the graph data, motifs with false correlations (such as motif6) will be connected in an adjacent manner in the test set, based on the set probability scenarios. However, when the probability conditions are not met, random connections will be established with graph data unrelated to the labels.
\item In the context of data quantity-based testing scenarios, we manifest variations in the quantity of spurious causal relationships by altering the multiplicative factor between confounding factors and genuine causal relationship data. Specifically, when a motif type associated with genuine causal relationships with labels emerges, varying quantities (reflecting the multiplicative factor) of spurious-associated motifs are connected in a manner that is distinct from the genuine relationship.
\item In the context of significance-based testing scenarios, we utilized the synthetically generated node classification dataset with rich features. Since the average and standard deviation of node features in this dataset can be configured, we opted to vary the standard deviation of generated nodes to manifest different levels of relationship complexity. The greater the standard deviation of nodes in the generated graph, the lower the significance of spurious causal relationship data.
\end{itemize}


\section{D. Settings and Details of Comparative Experiments}
\subsection{D.1. Baselines}
Consistent with the details outlined in Appendix C.1.
\subsection{D.2. Experimental Settings}
Experiments used a workstation with dual Quadro RTX 5000 GPUs (16 GB each), an Intel Xeon E5-1650 CPU, 128GB RAM, and Ubuntu 20.04 OS. Training involved Adam optimizer, 400 max epochs, and 32 batch size. SGD optimized Graph-SST5 and Graph-Twitter, while Spurious-Motif and CRCG utilized Gradient Descent (GD) for backpropagation.
\subsection{D.3. Dataset Details}
\subsubsection{Graph-SST5 dataset}
Graph-SST5\cite{dataset:sst} is a graph-structured dataset used for sentiment analysis, constructed based on the classic sentiment classification dataset SST5. The approach involves obtaining sentiment labels and sentence content, performing syntactic analysis to build syntactic parse trees, and transforming sentences into a graph structure. 

\subsubsection{Graph-Twitter dataset}
Graph-Twitter\cite{dataset:sst} is a social media text analysis dataset that collects Twitter tweets to construct user relationship graphs and tweet graphs. The user relationship graph is based on interaction behaviors, while the tweet graph is transformed into nodes representing vocabulary and edges representing semantics.
\subsubsection{Spurious-Motif dataset}
Spurious-Motif, re-implemented by \cite{DBLP:conf/iclr/WuWZ0C22}, has 18,000 graphs. Each has two subgraphs: one tied causally to the graph label, the other as a confounder. Both subgraph types have three variations. In training, a confounder subgraph is added with bias, e.g., 0.9 means 90\% class samples share it. Test sets randomly mix subgraphs, challenging causal understanding. Larger subgraphs intensify task complexity.
\subsubsection{CRCG}
CRCG is a dataset that we have generated ourselves.It provides a diverse and complex collection of graph datasets, facilitating thorough testing of GNNs in intricate scenarios. It not only encompasses a wide range of graph construction patterns but also allows pattern customization through parameter tuning. In our comparative experiments, we have chosen a representative scenario where spurious causal relationships appear in the training set with a probability of 20\%. This dataset design aims to capture the impact of causal factors on classification results, making it more aligned with real-world application scenarios.

\end{document}